\documentclass{article} % For LaTeX2e
\usepackage{iclr2021_conference,times}

\usepackage{amsmath,amsfonts,bm}

\def\eqref#1{equation~\ref{#1}}
\def\1{\bm{1}}

\DeclareMathAlphabet{\mathsfit}{\encodingdefault}{\sfdefault}{m}{sl}
\SetMathAlphabet{\mathsfit}{bold}{\encodingdefault}{\sfdefault}{bx}{n}

\usepackage{adjustbox}
\usepackage{algorithm}
\usepackage{algorithmic}
\usepackage{multirow}
\usepackage{subfig}
\usepackage{wrapfig}

\usepackage{pgfplots}
\pgfplotsset{compat = 1.3}

\renewcommand{\cite}[1]{\citep{#1}}

\usepackage{hyperref}
\usepackage{url}

\title{Optimizer Fusion: Efficient Training \\ with Better Locality and Parallelism}

\author{
Zixuan Jiang, Jiaqi Gu, Mingjie Liu, Keren Zhu, David Z. Pan \\
Electrical and Computer Engineering, The University of Texas at Austin \\
Austin, Texas 78712, USA \\
\texttt{\{zixuan, jqgu, jay\_liu, keren.zhu\}@utexas.edu, dpan@ece.utexas.edu}
}

\iclrfinalcopy % Uncomment for camera-ready version, but NOT for submission.
\begin{document}

\maketitle

\begin{abstract}
Machine learning frameworks adopt iterative optimizers to train neural networks.
Conventional eager execution separates the updating of trainable parameters from forward and backward computations.
However, this approach introduces nontrivial training time overhead due to the lack of data locality and computation parallelism. 
In this work, we propose to fuse the optimizer with forward or backward computation to better leverage locality and parallelism during training.
By reordering the forward computation, gradient calculation, and parameter updating, 
our proposed method improves the efficiency of iterative optimizers.
Experimental results demonstrate that we can achieve an up to $20\%$ training time reduction on various configurations. 
Since our methods do not alter the optimizer algorithm, they can be used as a general “plug-in” technique to the training process.
\end{abstract}

\section{Introduction}
\label{Introduction}

Iterative methods, such as stochastic gradient descent and its variants, are the mainstream optimization algorithms for training machine learning models.
In these methods, learnable parameters are updated step by step until the stopping criterion is met.
Many commonly used iterative optimization methods are implemented in popular machine learning frameworks,
e.g. PyTorch~\cite{pytorch}, TensorFlow~\cite{tensorflow}, MXNet~\cite{mxnet}, and Chainer~\cite{chainer}.

One critical component of these machine learning frameworks is automatic differentiation, which computes the gradients for all operations on tensors.
The smooth integration between the optimization kernel and automatic differentiation makes the training more accessible and boosts the popularization of these frameworks in the machine learning community.

Eager execution is widely adopted in these frameworks for its flexibility.
It usually decouples the forward propagation, gradient computation, and parameter updating into three separate stages.
In each iteration, forward computation is first performed.
Gradients respective to the loss function are then calculated for all learnable parameters.
Finally, learnable parameters are updated by a specified optimizer.
%Firstly, the learnable parameters are swept throughout the entire computational graph,
%and they cannot be efficiently reused in the local storage.
%Secondly, gradient updating operations are enforced to be after the gradient computation operations based on the implicit control dependency. 
%Further improving the data locality and algorithm parallelism can potentially increase training efficiency.
Although this implementation has an intuitive and transparent procedure, 
the learnable parameters and their gradients are read and written several times throughout one training iteration,
such that these data are not efficiently reused.
Moreover, gradient updating is enforced to be after the gradient computation based on the implicit control dependency,
leading to lower parallelism in the program execution.
In short, there is potential for higher training efficiency with better locality and parallelism in the eager execution.

In this work, we propose two methods \texttt{forward-fusion} and \texttt{backward-fusion},
which reorder the forward computation, gradient calculation, and parameter updating to accelerate the training process in eager execution.
Our proposed methods fuse the optimizer with forward or backward computation to better leverage locality and parallelism.
The \texttt{backward-fusion} method, 
motivated by the static computational graph compilation, where the optimizer is fused with gradient computation,
updates the parameters as early as possible.
For example, the optimizer in TensorFlow supports different gating gradients configurations.
\footnote{{\scriptsize \url{https://www.tensorflow.org/api_docs/python/tf/compat/v1/train/Optimizer##gating_gradients}}}
The \texttt{forward-fusion} method fuses the parameter updates with the next forward computation such that learnable parameters are updated as late as possible.
Like back-propagation through time (BPTT)~\cite{bptt}, the \texttt{forward-fusion} method expands the training process through iterations and uncovers a novel perspective of acceleration.

We summarize the advantages of our methods as follows.
\begin{itemize}
    \item Efficient. Our framework can increase the training speed by up to $20\%$.
    \item General. Our methods are orthogonal to other optimization methods and do not affect the training results.
    Thus, they can be applied in the training process of various machine learning tasks with different optimizers.
    We keep all the features of the eager execution.
    \item Simple. It is easy to replace the old training routines with our new methods.
    Users can accelerate their imperative training with little engineering effort.
\end{itemize}

\section{Background}
\label{background}

\begin{figure*}[ht]
    \centering
    \includegraphics[width=0.98\textwidth]{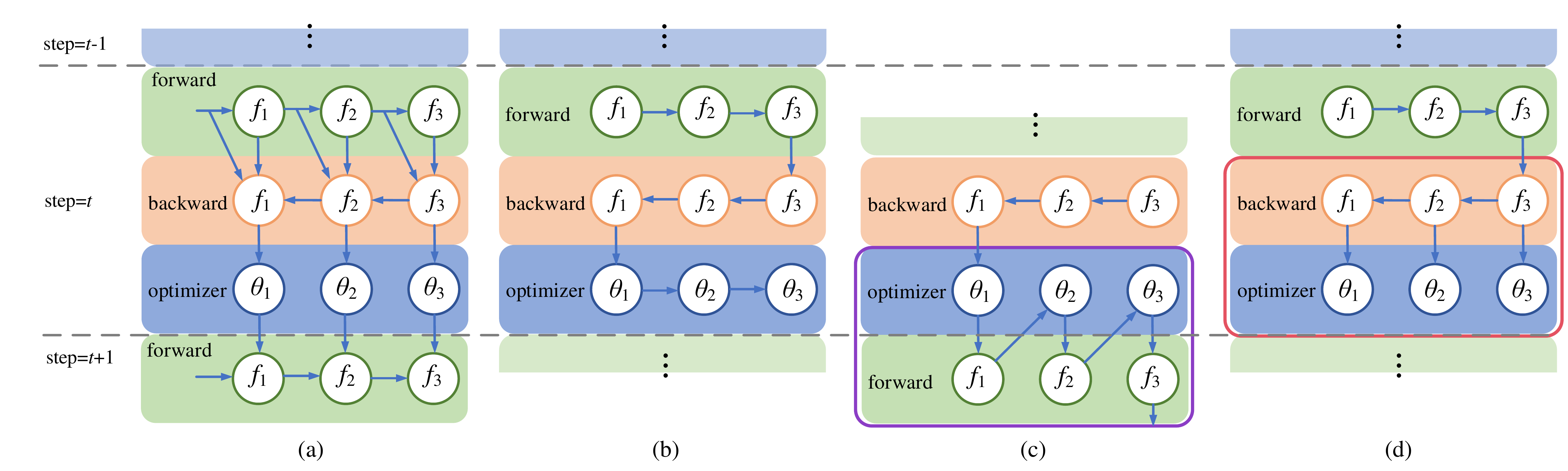}
    \caption{(a) Data dependency graph. (b) Baseline method. (c) \texttt{Forward-fusion}. (d) \texttt{Backward-fusion}.
    $\theta_i$ represents the trainable parameters in the layer $f_i$.}
    \label{fig:Framework}
\end{figure*}

\textbf{Static and Dynamic Computational Graphs.}
\label{background:computational_graph}
The training process can be viewed as a computational graph.
Figure~\ref{fig:Framework}(a) demonstrates the corresponding computation dependency for a three-layer neural network,
where nodes represent tensor computations, and directed edges stand for data dependencies.
Any topological order of this graph is a valid computation order.

Generally, there are two paradigms for tensor computations in machine learning frameworks.
The first category of framework compiles a model as a \emph{static (symbolic) computational graph} and executes the graph with all the neural network information.
For example, TensorFlow 1.X follows this routine by default~\cite{tensorflow}, which requires a pre-compiled graph before execution.
The other category of framework works in the \emph{eager (imperative) mode}, 
which immediately executes the newly-encountered computation node and incrementally builds a dynamic computational graph.
At each step, a computation node is appended to the current computational graph. 
PyTorch~\cite{pytorch} and TensorFlow 2.X~\cite{tensorflow} both run in eager mode by default,
enabling users to develop machine learning models more easily and quickly.
The eager mode also enables the efficient development of non-stationary neural architectures.
These two categories of frameworks are both widely used in the community.
For example, in the MLPerf training benchmark~\cite{mlperf}, both static and eager modes are used and achieve the state of the art performance.

\textbf{Graph Optimization.}
Engineers and practitioners in machine learning frameworks have proposed and implemented many \textit{graph optimizers},
such as TensorFlow's Grappler~\cite{grappler} and TASO~\cite{taso}.
The \texttt{backward-fusion} method has been considered in some of them.
However, to our best knowledge, existing graph optimizers need the information of the whole computation graph,
which means the optimizer fusion can only be used for (1) the static computation or (2) the mixture of static and dynamic execution.
In most machine learning frameworks, the purely eager mode still separates forward computation, backward pass, and parameter updating into three stages,
with the topological order shown in Figure~\ref{fig:Framework}(b).
The framework will first execute the forward and backward computation to obtain gradients. 
Then learnable parameters will be updated following the optimizer.
This imperative nature allows users to monitor the training process at the cost of low efficiency.
For instance, the TensorFlow eager execution follows this routine.~\footnote{\scriptsize \url{https://www.tensorflow.org/api_docs/python/tf/compat/v1/train/Optimizer##minimize}}

We enable the \texttt{backward-fusion} method in purely eager mode.
The \texttt{forward-fusion} method also uncovers a novel perspective, where graphs can be optimized across iterations.

\begin{wrapfigure}{R}{0.5\textwidth}
    \centering
    \includegraphics[width=0.5\textwidth]{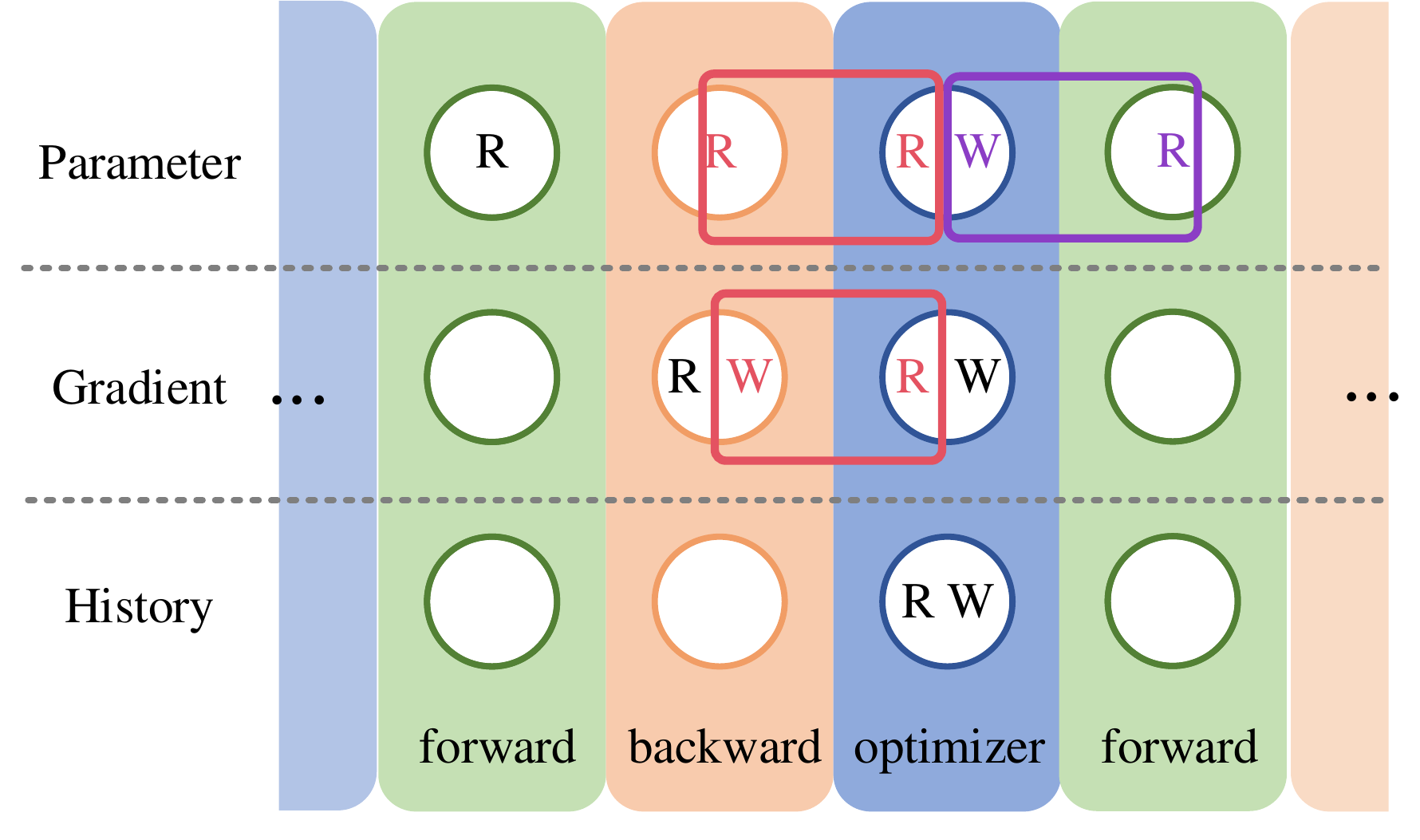}
    \caption{Memory transactions and data locality in the training process. \textit{R} and \textit{W} represent memory read and write, respectively. \textit{History} means parameter history needed in the optimizer, e.g., momentum. Red and purple frames represent locality improvement in \texttt{backward-fusion} and \texttt{forward-fusion}, respectively.}
    \label{fig:DataLocality}
\end{wrapfigure}

% \begin{figure}[ht]
%     \centering
%     \includegraphics[width=0.45\textwidth]{figure/Locality.pdf}
%     \caption{Memory transactions and data locality in the training process. \textit{R} and \textit{W} represent memory read and write, respectively. \textit{History} means parameter history needed in the optimizer, e.g., momentum. Red and purple frames represent locality improvement in \texttt{backward-fusion} and \texttt{forward-fusion}, respectively.}
%     \label{fig:DataLocality}
% \end{figure}

\section{Methods}
\label{methods}
Figure~\ref{fig:Framework} shows the baseline method, and the proposed forward-fusion and backward-fusion methods.

\textbf{Baseline.}
Figure~\ref{fig:DataLocality} illustrates the memory transactions and the locality that can be leveraged in the training process.
Trainable parameters are read during forward and backward computations and updated by the optimizer.
Gradients are accumulated during the backward pass. Finally, they are read and reset by the optimizer.
History represents the parameter history, such as momentum.
The optimizer records and updates the parameter history.

All these memory reads and writes are separated by forward, backward, and optimizer stages.
The memory capacity is usually not large enough to hold all the data through the training iteration.
The data locality between the optimizer and its adjacent forward or backward computations is lost.
If we access the same set of data repetitively before the data is flushed, we can shorten the time of memory access and thus accelerate the training process.

Also, the baseline method does not take advantage of the parallelism between backward computations and parameter updating.
While updating a group of parameters,
we can continue the back-propagation to compute gradients for other independent parameters at the same time,
which offers another opportunity for better parallelism in the training process. 

\textbf{Forward-Fusion.}
One approach is to fuse the optimizer with the forward computation in the next iteration.
The next forward pass can occur in either a training or an evaluation process.
Each trainable parameter will be updated as late as possible.
This \textit{lazy} update strategy is named \texttt{forward-fusion}.
The proposed method can be applied in all the iterative methods, including the optimizer that needs global information.

The memory write operation during parameter updating can be merged with the next read, 
such that in the next forward computation,
the cached parameter can be quickly accessed with low latency.
The purple frame in Figure~\ref{fig:DataLocality} illustrates this improvement.

Unlike current machine learning frameworks, which ignore the optimization potential across different iterations,
we find the opportunity between adjacent training steps.
Similar to back-propagation through time~\cite{bptt},
the training process of a feed-forward neural network training can also be expanded through iterations.
This perspective is a new direction for accelerating the machine learning models embracing both static and dynamic computational graphs.

\textbf{Backward-Fusion.}
Another approach is to fuse the optimization computation with gradient computation in the backward pass, as shown in Figure~\ref{fig:Framework}(d).
This method applies gradients to parameters as early as possible,
so that the memory access can be merged to increase the locality,
as shown in the red frame in Figure~\ref{fig:DataLocality}.
Specifically, two consecutive parameter reads in the backward pass and optimizer can be merged,
such that the second read during the optimization step can be accelerated as it can be cached in the local storage.
Gradient accumulation in the backward computation can be merged with the memory read in the optimizer.
Thus, the updated gradients can be efficiently accessed in the local storage to shorten the memory access time.

Moreover, this method improves training efficiency as it parallelizes the parameter updating and gradient back-propagation.
This method applies to most optimizers that do not require global information of trainable parameters,
as this fusion strategy assumes the update of $\theta_i$ is decoupled with other parameters $\theta_j (i \neq j)$.
In the view of the computational graph, 
the depths of the directed graphs shown in Figures~\ref{fig:Framework}(b), (d) are $3n$ and $2n+1$, respectively, where $n$ is the number of layers in the neural network.
Thus, the \texttt{backward-fusion} method also provides extra parallelism.

Table~\ref{table:LocalityComparison} summarizes the characteristics of our proposed methods.

\begin{table}[ht]
\caption{Comparison among three methods: locality, parallelism, and global information.}
\small
\centering
\label{table:LocalityComparison}
\begin{adjustbox}{width=0.48\textwidth,center}
\begin{tabular}{cccc} \hline \hline
Method          & Locality     & Parallelism  & Global Info. \\ \hline
baseline        &   $\times$   & $\times$     & $\checkmark$ \\
forward-fusion  & $\checkmark$ & $\times$     & $\checkmark$ \\
backward-fusion & $\checkmark$ & $\checkmark$ & $\times$ \\
\hline \hline          
\end{tabular}
\end{adjustbox}
\end{table}

\section{Experiments}
\label{experiments}

\begin{figure}[ht]
    \centering
\begin{tikzpicture}
\begin{axis}[
    width=0.45\textwidth,
    every node near coord/.style={
        /pgf/number format/fixed,
        /pgf/number format/fixed zerofill,
        /pgf/number format/precision=2
    },
    ybar stacked,
	bar width=15pt,
	nodes near coords,
    enlargelimits=0.15,
    legend style={at={(0.5,-0.20)},
      anchor=north,legend columns=-1},
    ylabel={Time per iteration (ms)},
    symbolic x coords={baseline, FF, BF},
    xtick=data,
    %x tick label style={rotate=0, anchor=east},
    ]
\addplot+[ybar] plot coordinates {(baseline,21.90) (FF, 28.65) (BF, 21.99)};
\addplot+[ybar] plot coordinates {(baseline,55.61) (FF, 55.77) (BF, 58.93)};
\addplot+[ybar] plot coordinates {(baseline,16.70) (FF, 0) (BF, 0)};
\legend{\strut forward, \strut backward, \strut optimzer}
\end{axis}
\end{tikzpicture}
    \caption{Training time breakdown of MobileNetV2 with mini-batch size 32.
    FF, BF are short for forward-fusion, backward-fusion.}
    \label{fig:runtime_breakdown}
\end{figure}
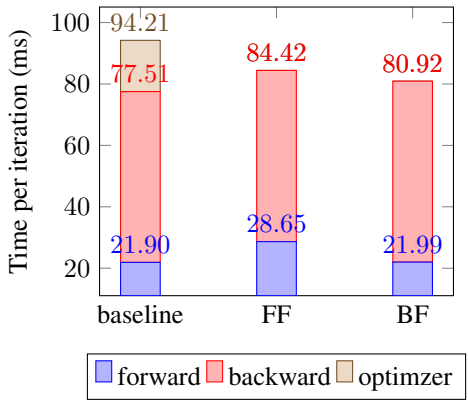

Figure~\ref{fig:runtime_breakdown} shows the time breakdown of one training iteration of MobileNetV2~\cite{mobilenetv2} with a mini-batch size of 32.
After we fuse the optimizer with backward computation,
the execution time of backward increases by $3.32$ ms,
much smaller than the original optimizer execution time ($16.70$ ms).
In this example, our \texttt{forward-fusion} and \texttt{backward-fusion} improve the training throughput by $12\%$ and $16\%$, respectively.

We show more results in Appendix~\ref{appendix:experiment}.

\section{Conclusions}
\label{conclusion}

Conventional eager execution in machine learning frameworks
separate the updating of trainable parameters from forward and backward computations.
In this paper, we propose two methods \texttt{forward-fusion} and \texttt{backward-fusion}
to better leverage the locality and parallelism during training.
We reorder the forward computation, gradient calculation, and parameter updating, 
so that our proposed methods improve the efficiency of iterative optimizers.
Experimental results demonstrate the effectiveness and efficiency of our methods across various configurations.
Our \texttt{forward-fusion} method opens a new perspective of performance optimization for machine learning frameworks.

% For future directions, we plan to extend our methods to distributed training,
% where there is parallelism across different machines.
% The gradient computation and parameter updating can be merged to save time.
% Also, we intend to refactor our implementation since our methods are currently implemented on the frontend.

\bibliography{iclr2021_conference}
\bibliographystyle{iclr2021_conference}

\appendix
\section{Extended Related Work}

\textbf{Iterative Optimization Methods.}
\label{background:iterative}
An iterative optimization algorithm starts from an initial guess and derives a sequence of improving approximate solutions. 
Algorithm~\ref{alg:optimization_algorithms} is the general structure of iterative optimization methods for unconstrained problems.
The step vector $\Delta \theta$ is computed from the optimizer policy $\pi$.
The policy $\pi$ is the only difference across different optimization algorithms.
The commonly used gradient-based methods use policies that depend on the first-order derivative information.

For instance, the policies demonstrated in Equations~(\ref{eq:gd}), (\ref{eq:momentum}), and~(\ref{eq:newton})
are used in the gradient descent, gradient descent with momentum, Newton's method, respectively,
\begin{equation}
   \label{eq:gd}
   \pi_1 = -\eta \nabla f(\theta^{(t - 1)})  
\end{equation}
\begin{equation}
    \label{eq:momentum}
    \pi_2 = -\eta \sum_{\tau=0}^{t-1} \alpha^{t-\tau-1} \nabla f(\theta^{(\tau)})
\end{equation}
\begin{equation}
    \label{eq:newton}
    \pi_3 = -\eta \nabla ^2f(\theta^{(t - 1)})^{-1} \nabla f(\theta^{(t - 1)})
\end{equation}
where $\eta$ represents the step size, $\alpha$ denotes the momentum decay factor.

\begin{algorithm}[ht]
   \caption{Optimization algorithms in general form}
   \label{alg:optimization_algorithms}
\begin{algorithmic}
   \STATE {\bfseries Input:} objective function $f$
   \STATE Initialize the starting point $\theta^{(0)}$ 
   \FOR{$t = 1, 2, ...$}
        \IF{stopping criterion is met}
            \STATE \textbf{return} $\theta^{(t-1)}$
        \ENDIF
        
        \STATE $\Delta \theta = \pi(f, {\theta^{(0)}, \theta^{(1)}, ..., \theta^{(t-1)}})$
        \STATE $\theta^{(t)} = \theta^{(t-1)} + \Delta \theta$
   \ENDFOR
\end{algorithmic}
\end{algorithm}

\textbf{Locality and Parallelism.}
\label{background:locality_parallelism}
Fruitful hardware-aware techniques in machine learning have been proposed to accelerate the training process, especially on graphics processing units (GPUs).
Generally speaking, common approaches include 
accelerating the kernel computations~\cite{efficient_gemm}, 
mixed precision training~\cite{mixed_precision_training} ,
fusing kernels and operations~\cite{kernel_fusion},
and exploring efficient network architecture~\cite{mixed_precision_nas}.
Data locality and computation parallelism are two critical aspects of performance optimization.

The work of fused-layer CNN accelerators~\cite{MICRO_2016_Alwani} proposes a new architecture for inference by fusing the convolution layers. 
It decomposes the inputs to the convolution layers into tiles and propagates one tile through multiple layers.
With reduced memory access and better cache utilization, the inference speed is improved.
% However, the approach is nontrivial to be applied in training.
Lym et al.~\cite{Arxiv_2018_Lym} design a new scheme to eliminate most memory accesses in neural network training by reordering the computation within a mini-batch for better data locality.
Apex for PyTorch~\footnote{\url{https://github.com/NVIDIA/apex}} uses fused optimizers, which launch one kernel for the element-wise operations.

From a perspective different from the aforementioned previous works,
we explore the parallelism and locality across the gradient computation and the optimizer.
We also uncover the potential of acceleration across iterations.

\section{Extended Method}

\subsection{Forward-fusion}

Algorithm~\ref{alg:fusion_forward} shows the pipeline of this method.
It is possible that a layer is used many times, i.e., $f_i = f_j, i \neq j$ in Algorithm ~\ref{alg:fusion_forward}.
We use a flag \texttt{updated} to ensure that the parameter is updated only once no matter how many times the corresponding layer is used.
This method can also be applied when we need to manipulate the gradients based on global information~\cite{amsgrad}.
For instance, this method is suitable when we would like to clip the gradients by their global norm.

\begin{algorithm}[ht]
   \caption{Forward-fusion}
   \label{alg:fusion_forward}
\begin{algorithmic}
    \STATE {\bfseries Input:} topologically sorted operator array $\{f_i\}_{i=1}^n$
    \FOR{$i = 1, 2, ..., n$}
        \IF{$f_i$.updated is False}
            \FOR{each trainable parameters $\theta$ in $f_i$}
                \STATE execute optimizer of $\theta$
            \ENDFOR
            \STATE $f_i$.updated $\gets$ True
        \ENDIF
        \STATE execute $f_i$
    \ENDFOR
    \FOR{$i = n, n - 1, ..., 1$}
        \STATE execute backward pass of $f_i$
        \STATE accumulate gradients for all trainable parameters
        \STATE $f_i$.updated $\gets$ False
    \ENDFOR
\end{algorithmic}
\end{algorithm}

\subsection{Backward-fusion}

Algorithm~\ref{alg:fusion_backward} demonstrates the computational flow.
After calculating the gradient of $\theta_i$, we apply the gradient in the optimizer directly.
At the same time, we resume the backward computation for node $f_{i-1}$.
For each parameter $\theta_i$, we record the number of its usage in the forward pass as $\theta_i$.count.
Correspondingly, we will update it until its gradient is accumulated for all its usage in the forward pass.

\begin{algorithm}[ht]
   \caption{Backward-fusion}
   \label{alg:fusion_backward}
\begin{algorithmic}
    \STATE {\bfseries Input:} topologically sorted operator array $\{f_i\}_{i=1}^n$
    \FOR{$i = 1, 2, ..., n$}
        \STATE execute $f_i$
        \FOR{each trainable parameters $\theta$ in $f_i$}
            \STATE $\theta$.count $\gets$ $\theta$.count $+1$
        \ENDFOR
    \ENDFOR
    \FOR{$i = n, n - 1, ..., 1$}
        \STATE execute backward pass of $f_i$
        \FOR{each trainable parameters $\theta$ in $f_i$}
            \STATE accumulate gradient of $\theta$
            \STATE $\theta$.count $\gets$
            $\theta$.count $-1$
            \IF{$\theta$.count is 0}
                \STATE execute optimizer of $\theta$
            \ENDIF
        \ENDFOR
    \ENDFOR
\end{algorithmic}
\end{algorithm}

Applying gradients directly on the parameters may induce race conditions.
For instance, for a multiplication operator $f(\theta, x) = \theta x$, $\partial f / \partial \theta = x, \partial f / \partial x = \theta$.
$\partial f / \partial x$ depends on the original parameter $\theta^{(t)}$, instead of the updated parameter $\theta^{(t+1)}$.
Therefore, we must carefully tackle this dependency.
Specifically, for a trainable parameter $\theta$, we will update it in-place to obtain $\theta^{(t+1)}$
when the following two conditions are both satisfied:
(1) its gradient $\partial L / \partial \theta$ is calculated, 
and (2) there is no other dependency on the old value $\theta^{(t)}$.

\section{Experiment details}
\label{appendix:experiment}

We evaluate the effectiveness and efficiency of our proposed methods with various 
mini-batch sizes, 
optimizers, 
machine learning models and benchmarks,
machines (GPUs) 
and frameworks.

\subsection{Experimental Settings}
Unless stated otherwise, we conduct experiments on the eager execution in PyTorch 1.6.0.
We implement the proposed methods in the PyTorch front-end using hooks.
A toy example is provided in the supplementary file.
The training process runs on a Linux server with Intel Core i9-7900X CPU and a NVIDIA TITAN Xp GPU based on Pascal architecture.
We use Adam~\cite{adam} with weight decay to do the training on image classification using the ImageNet dataset~\cite{imagenet}.
All the tensor computations occur on 1 GPU with single-precision floating-point (float32) datatype.

We report the mean of 100 training iterations.

\subsection{Various Mini-batch Sizes}

\begin{figure}[ht]
    \centering
    \begin{tikzpicture}
        \begin{axis}[
            width=0.45\textwidth,
            height=5cm,
            ymajorgrids,
            xlabel = {mini-batch size},
            ylabel={saved time (ms)},
            x tick label style={/pgf/number format/1000 sep=},
            ytick scale label code/.code={$\times$bn},
            legend pos= south east
        ]
        \addplot[blue]
            table[x=idx,y=forward-fusion] {saved_time_mobilenetv2.csv};
        \addlegendentry{forward-fusion}
        \addplot[red]
            table[x=idx,y=backward-fusion] {saved_time_mobilenetv2.csv};
        \addlegendentry{backward-fusion}
        \end{axis}
    \end{tikzpicture}
    \caption{The absolute execution time saved by our methods on MobileNetV2 with different mini-batch sizes.}
    \label{fig:absolute_time_saving}
\end{figure}
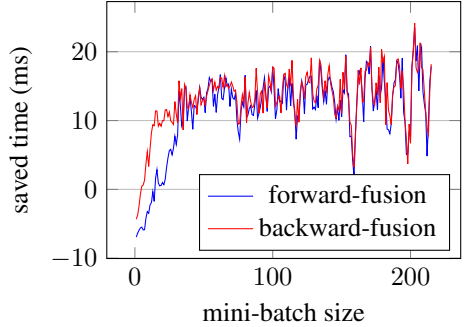

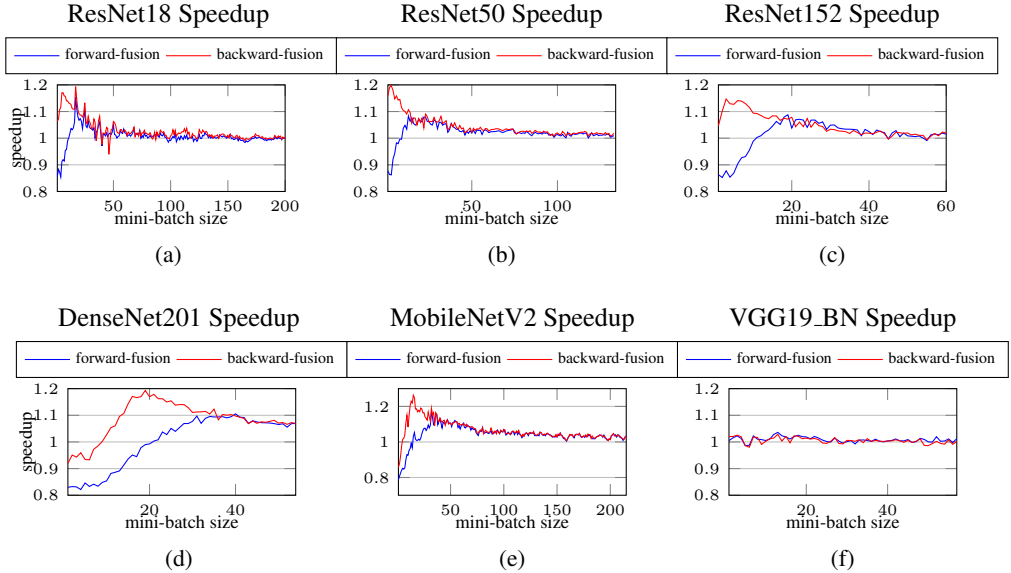
\begin{figure*}[ht]
    \centering
    \subfloat[]{
    \noindent\begin{tikzpicture}
        \begin{axis}[
            width=0.33\textwidth,
            height=3cm,
            every node/.style={inner sep=0,outer sep=0},
            ymajorgrids,
            xlabel = {mini-batch size},
            ylabel={speedup},
            x label style={at={(axis description cs:0.5,-0.1)},anchor=north},
    y label style={at={(axis description cs:-0.1,.5)},rotate=0,anchor=south},
            label style={font=\scriptsize},
            x tick label style={font = \tiny, /pgf/number format/1000 sep=},
            y tick label style={font = \tiny, /pgf/number format/1000 sep=},
            ytick scale label code/.code={$\times$bn},
            legend columns=-1,
            legend style={at={(0.5,1.1)},anchor=south, font=\tiny},
            ymin=0.8,
            ymax=1.2,
            xmin=1,
            xmax=200
        ]
        \addplot[blue]
            table[x=idx,y=forward-fusion] {resnet18_speedup.csv};
        \addlegendentry{forward-fusion}
        \addplot[red]
            table[x=idx,y=backward-fusion] {resnet18_speedup.csv};
        \addlegendentry{backward-fusion}
        \end{axis}
        \node[above] at (current bounding box.north) {ResNet18 Speedup};
    \end{tikzpicture}
    \label{fig:ResNet18Sppedup}
    }
    \hspace{-8pt}
    \subfloat[]{
    \noindent\begin{tikzpicture}
        \begin{axis}[
            width=0.33\textwidth,
            height=3cm,
            every node/.style={inner sep=0,outer sep=0},
            ymajorgrids,
            xlabel = {mini-batch size},
            x label style={at={(axis description cs:0.5,-0.1)},anchor=north},
    y label style={at={(axis description cs:-0.1,.5)},rotate=0,anchor=south},
            label style={font=\scriptsize},
            x tick label style={font = \tiny, /pgf/number format/1000 sep=},
            y tick label style={font = \tiny, /pgf/number format/1000 sep=},
            ytick scale label code/.code={$\times$bn},
            legend columns=-1,
            legend style={at={(0.5,1.1)},anchor=south, font=\tiny},
            ymin=0.8,
            ymax=1.2,
            xmin=1,
            xmax=134
        ]
        \addplot[blue]
            table[x=idx,y=forward-fusion] {resnet50_speedup.csv};
        \addlegendentry{forward-fusion}
        \addplot[red]
            table[x=idx,y=backward-fusion] {resnet50_speedup.csv};
        \addlegendentry{backward-fusion}
        \end{axis}
        \node[above] at (current bounding box.north) {ResNet50 Speedup};
    \end{tikzpicture}
    \label{fig:ResNet50Sppedup}
    }\hspace{-8pt}
    \subfloat[]{
    \noindent\begin{tikzpicture}
        \begin{axis}[
            width=0.33\textwidth,
            height=3cm,
            every node/.style={inner sep=0,outer sep=0},
            ymajorgrids,
            xlabel = {mini-batch size},
            x label style={at={(axis description cs:0.5,-0.1)},anchor=north},
    y label style={at={(axis description cs:-0.1,.5)},rotate=0,anchor=south},
            label style={font=\scriptsize},
            x tick label style={font = \tiny, /pgf/number format/1000 sep=},
            y tick label style={font = \tiny, /pgf/number format/1000 sep=},
            ytick scale label code/.code={$\times$bn},
            legend columns=-1,
            legend style={at={(0.5,1.1)},anchor=south, font=\tiny},
            ymin=0.8,
            ymax=1.2,
            xmin=1,
            xmax=60
        ]
        \addplot[blue]
            table[x=idx,y=forward-fusion] {resnet152_speedup.csv};
        \addlegendentry{forward-fusion}
        \addplot[red]
            table[x=idx,y=backward-fusion] {resnet152_speedup.csv};
        \addlegendentry{backward-fusion}
        \end{axis}
        \node[above] at (current bounding box.north) {ResNet152 Speedup};
    \end{tikzpicture}
    \label{fig:ResNet152Sppedup}
    }\hspace{-8pt}
    \\
    \subfloat[]{
    \noindent\begin{tikzpicture}
        \begin{axis}[
            width=0.33\textwidth,
            height=3cm,
            every node/.style={inner sep=0,outer sep=0},
            ymajorgrids,
            xlabel = {mini-batch size},
            ylabel={speedup},
            x label style={at={(axis description cs:0.5,-0.1)},anchor=north},
    y label style={at={(axis description cs:-0.1,.5)},rotate=0,anchor=south},
            label style={font=\scriptsize},
            x tick label style={font = \tiny, /pgf/number format/1000 sep=},
            y tick label style={font = \tiny, /pgf/number format/1000 sep=},
            ytick scale label code/.code={$\times$bn},
            legend columns=-1,
            legend style={at={(0.5,1.1)},anchor=south, font=\tiny},
            ymin=0.8,
            ymax=1.2,
            xmin=1,
            xmax=54
        ]
        \addplot[blue]
            table[x=idx,y=forward-fusion] {densenet201_speedup.csv};
        \addlegendentry{forward-fusion}
        \addplot[red]
            table[x=idx,y=backward-fusion] {densenet201_speedup.csv};
        \addlegendentry{backward-fusion}
        \end{axis}
        \node[above] at (current bounding box.north) {DenseNet201 Speedup};
    \end{tikzpicture}
    \label{fig:DenseNet201Sppedup}
    }\hspace{-8pt}
     \subfloat[]{
    \noindent\begin{tikzpicture}
        \begin{axis}[
            width=0.33\textwidth,
            height=3cm,
            every node/.style={inner sep=0,outer sep=0},
            ymajorgrids,
            xlabel = {mini-batch size},
            x label style={at={(axis description cs:0.5,-0.1)},anchor=north},
    y label style={at={(axis description cs:-0.1,.5)},rotate=0,anchor=south},
            label style={font=\scriptsize},
            x tick label style={font = \tiny, /pgf/number format/1000 sep=},
            y tick label style={font = \tiny, /pgf/number format/1000 sep=},
            ytick scale label code/.code={$\times$bn},
            legend columns=-1,
            legend style={at={(0.5,1.1)},anchor=south, font=\tiny},
            ymin=0.7,
            ymax=1.3,
            xmin=1,
            xmax=215
        ]
        \addplot[blue]
            table[x=idx,y=forward-fusion] {mobilenetv2_speedup.csv};
        \addlegendentry{forward-fusion}
        \addplot[red]
            table[x=idx,y=backward-fusion] {mobilenetv2_speedup.csv};
        \addlegendentry{backward-fusion}
        \end{axis}
        \node[above] at (current bounding box.north) {MobileNetV2 Speedup};
    \end{tikzpicture}
    \label{fig:MobileNetV2Sppedup}
    }\hspace{-8pt}
     \subfloat[]{
    \noindent\begin{tikzpicture}
        \begin{axis}[
            width=0.33\textwidth,
            height=3cm,
            every node/.style={inner sep=0,outer sep=0},
            ymajorgrids,
            xlabel = {mini-batch size},
            x label style={at={(axis description cs:0.5,-0.1)},anchor=north},
    y label style={at={(axis description cs:-0.1,.5)},rotate=0,anchor=south},
            label style={font=\scriptsize},
            x tick label style={font = \tiny, /pgf/number format/1000 sep=},
            y tick label style={font = \tiny, /pgf/number format/1000 sep=},
            ytick scale label code/.code={$\times$bn},
            legend columns=-1,
            legend style={at={(0.5,1.1)},anchor=south, font=\tiny},
            ymin=0.8,
            ymax=1.2,
            xmin=1,
            xmax=57
        ]
        \addplot[blue]
            table[x=idx,y=forward-fusion] {vgg19_bn_speedup.csv};
        \addlegendentry{forward-fusion}
        \addplot[red]
            table[x=idx,y=backward-fusion] {vgg19_bn_speedup.csv};
        \addlegendentry{backward-fusion}
        \end{axis}
        \node[above] at (current bounding box.north) {VGG19\_BN Speedup};
    \end{tikzpicture}
    \label{fig:VGG19Sppedup}
    }\hspace{-8pt}
    
    \caption{Training speedup with various mini-batch sizes on different benchmarks.}
    \vspace{-8pt}
    \label{fig:SpeedupPlots}
\end{figure*}

%\begin{figure}[ht]
%\begin{center}
%\centerline{\includegraphics[width=\columnwidth]{figure/saved_time.pdf}}
%\caption{The absolute runtime saved by our methods on MobileNetV2.}
%\label{fig:absolute_time_saving}
%\end{center}
%\end{figure}

%\begin{figure*}[tb]
%    \centering
%    \subfloat[]{
%    \includegraphics[width=0.33\textwidth]{figure/resnet18.pdf}
%    \label{fig:ResNet18Sppedup}
%    }
%    \hspace{-8pt}
%    \subfloat[]{
%    \includegraphics[width=0.33\textwidth]{figure/resnet50.pdf}
%    \label{fig:ResNet50Sppedup}
%    }\hspace{-8pt}
%    \subfloat[]{
%    \includegraphics[width=0.33\textwidth]{figure/resnet152.pdf}
%    \label{fig:ResNet152Sppedup}
%    }\hspace{-8pt}
%    \\
%    \subfloat[]{
%    \includegraphics[width=0.33\textwidth]{figure/densenet201.pdf}
%    \label{fig:DenseNet201Sppedup}
%    }\hspace{-8pt}
%     \subfloat[]{
%    \includegraphics[width=0.33\textwidth]{figure/mobilenet.pdf}
%    \label{fig:MobileNetV2Sppedup}
%    }\hspace{-8pt}
%     \subfloat[]{
%    \includegraphics[width=0.33\textwidth]{figure/vgg19bn.pdf}
%    \label{fig:VGG19Sppedup}
%    }\hspace{-8pt}
%    
%    \caption{Training speedup with various mini-batch sizes on different benchmarks.}
%    \vspace{-8pt}
%    \label{fig:SpeedupPlots}
%\end{figure*}

Compared with the baseline method, our two methods have the overhead of control as shown in Algorithms~\ref{alg:fusion_forward} and~\ref{alg:fusion_backward}.
When a small mini-batch size is used, 
the overhead of the control flow will exceed the benefits of better locality and parallelism,
which makes our framework slower than the baseline.
As the mini-batch size increases,
the overhead becomes negligible compared with the computation time.
The effectiveness of our proposed methods requires this overhead to be amortized by an appropriate mini-batch size.

When the mini-batch size is large enough such that we reach the performance roofline of the GPUs,
the computation time of forward and backward pass is approximately linear to the mini-batch size.
On the other hand, the optimizer execution time is independent of the mini-batch size.
Therefore, the absolute training time saved by our methods is independent of the mini-batch size, 
as shown in Figure~\ref{fig:absolute_time_saving}.
The relative speedup will decrease as the mini-batch size grows, as demonstrated in Figure~\ref{fig:SpeedupPlots}.

The discussion above can also be formulated in the following equation.
The theoretical speedup of the training process is
$$
s = \frac{bt_{grad} + t_{opt}}{bt_{grad} + t_{opt} - t_{saved}}
$$
where $b$ represents the mini-batch size,
$t_{grad}$ is the time of forward and backward computation per mini-batch size,
$t_{opt}$ is the execution time of optimizer,
and $t_{saved}$ stands for the absolute saved time on the optimizer with our methods.

Both \texttt{forward-fusion} and \texttt{backward-fusion} methods leverage the locality of the trainable parameters.
However, only the \texttt{backward-fusion} method takes advantage of the parallelism between the gradient computation and optimizer.
When the mini-batch size is small, the GPU is not fully utilized.
Thus, the parallelism exploited by \texttt{backward-fusion} will accelerate the training significantly compared with \texttt{forward-fusion}.
% That is the reason why \texttt{backward-fusion} is faster than the \texttt{forward-fusion} when mini-batch size is small.
As mini-batch sizes grow, the gradient computation dominates the GPU utilization.
%The parallelism in the \texttt{backward-fusion} can not be taken efficiently.
Therefore, the execution time of these two methods converges at large mini-batch sizes, as illustrated in Figure~\ref{fig:SpeedupPlots}.

\subsection{Various Models and Optimizers}

\begin{figure}[ht]
\centering
    \begin{tikzpicture}
        \begin{axis}[
            width=0.5\textwidth,
            height=6cm,
            xmode=log,
            log ticks with fixed point,
            ymajorgrids,
            xlabel = {average \#parameter per layer ($10^3$)},
            ylabel={speedup},
            x tick label style={/pgf/number format/1000 sep=},
            ytick scale label code/.code={$\times$bn},
            legend columns=-1,
            legend style={at={(0.5,1.1)},anchor=south, font=\tiny}
        ]
        \addplot[
            scatter,only marks,scatter src=explicit symbolic,
            scatter/classes={
                a={mark=square*,red},
                b={mark=square*,blue},
                c={mark=triangle*,red},
                d={mark=triangle*,blue},
                e={mark=o,red},
                f={mark=o,blue}
            },
            nodes near coords*={\Label},
            every node near coord/.append style={xshift=0pt,yshift=0pt,anchor=south,font=\tiny},
            visualization depends on={value \thisrow{net} \as \Label}
        ]
            table[x=param_size,y=speedup,meta=model] {model_speedup.csv};
            % \legend{ResNet18, ResNet50, ResNet152, DenseNet201, MobileNetV2, VGG19\_bn}
        \end{axis}
    \end{tikzpicture}
\caption{The speedup trend among different models with a mini-batch size of 32.
In average, fewer parameters per layer leads to higher speedup.}
\label{fig:model_comparison}
\end{figure}
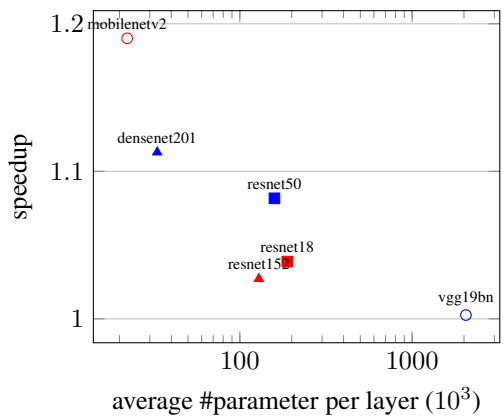

We sweep the mini-batch size for different models~\cite{resnet, densenet, mobilenetv2, vgg, batch_norm} as shown in Figure~\ref{fig:SpeedupPlots}.
Figure~\ref{fig:model_comparison} demonstrates the relationship between the parameter size and speedup across different models.
The smaller the average number of parameters per layer, the more locality we can leverage
so that our methods can achieve higher training speed.
This explains why the VGG19\_BN is hardly accelerated while the MobileNetV2 has the most significant improvement.
Currently, the models targeting at edge devices usually contain fewer parameters,
whose training will benefit more from our methods.

%\begin{figure}[ht]
%\begin{center}
%\centerline{\includegraphics[width=\columnwidth]{figure/bubble.pdf}}
%\caption{The speedup trend among different models with the mini-batch size of 32.
%In average, fewer parameters per layer leads to higher speedup.}
%\label{fig:model_comparison}
%\end{center}
%\end{figure}

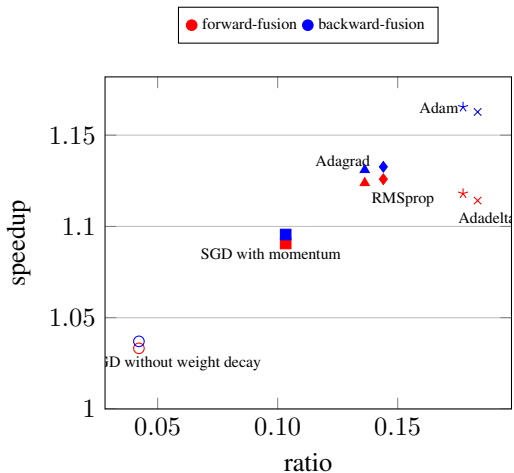
\begin{figure}[htbp]
\begin{center}
\centering
    \begin{tikzpicture}
        \begin{axis}[
            width=0.5\textwidth,
            height=6cm,
            ymajorgrids,
            xlabel = {ratio},
            ylabel = {speedup},
            ymin = 1,
            x tick label style={
                        /pgf/number format/fixed,
            /pgf/number format/precision=2,
            /pgf/number format/fixed zerofill
        },
        scaled x ticks=false,
            ytick scale label code/.code={$\times$bn},
            legend entries={forward-fusion, backward-fusion},%, SGD without weight decay, SGD with momentum,
            %Adagrad, RMSprop, Adam, Adadelta},
            legend columns=2,
            legend style={at={(0.5,1.1)},anchor=south, font=\tiny}
        ]
        \addlegendimage{only marks, mark=*, red}
        \addlegendimage{only marks, mark=*, blue}
        %\addlegendimage{only marks, mark=o}
        %\addlegendimage{only marks, mark=square*}
        %\addlegendimage{only marks, mark=triangle*}
        %\addlegendimage{only marks, mark=diamond*}
        %\addlegendimage{only marks, mark=star}
        %\addlegendimage{only marks, mark=x}
        \addplot[
            scatter,only marks,scatter src=explicit symbolic,
            scatter/classes={
                1={mark=o,red},
                2={mark=square*,red},
                3={mark=triangle*,red},
                4={mark=diamond*,red},
                5={mark=star,red},
                6={mark=x,red},
                7={mark=o,blue},
                8={mark=square*,blue},
                9={mark=triangle*,blue},
                10={mark=diamond*,blue},
                11={mark=star,blue},
                12={mark=x,blue}
            },
        ]
            table[x=ratio,y=speedup,meta=type] {optimizer_speedup.csv};
            \node[font=\tiny] at (15,25) {SGD without weight decay};
            \node[font=\tiny] at (55,85) {SGD with momentum};
            \node[font=\tiny] at (85,135) {Adagrad};
            \node[font=\tiny] at (110,115) {RMSprop};
            \node[font=\tiny] at (125,165) {Adam};
            \node[font=\tiny] at (145,105) {Adadelta};
        \end{axis}
    \end{tikzpicture}
\caption{Comparison among various optimizers on MobileNetV2 with a mini-batch size of 32.
Weight decay is applied in all these methods unless specified.
The horizontal axis represents the ratio of the optimizer time to a whole iteration time.}
\label{fig:various_optimizer}
\end{center}
\end{figure}

%\begin{figure}[h]
%\begin{center}
%\centerline{\includegraphics[width=\columnwidth]{figure/various_optimizer.pdf}}
%\caption{Comparison among various optimizers on MobileNetV2 with a mini-batch size of 32.
%Weight decay is applied in all these methods unless specified.
%The horizontal axis represents the ratio of the optimizer time to a whole iteration time.}
%\label{fig:various_optimizer}
%\end{center}
%\end{figure}

Various optimizers used in machine learning frameworks can benefit from our proposed methods.
Figure~\ref{fig:various_optimizer} shows an increasing trend between speedup and the runtime ratio of different optimizers~\cite{adadelta, adam, adagrad}.
The horizontal axis is the ratio of the optimizer runtime to a whole iteration runtime.
The more runtime-costly the optimizer, the higher speedup we can achieve.

\subsection{Various Machines and Benchmarks}

\begin{table*}[ht]
\caption{Training results on MobileNetV2 with a mini-batch size of 32 across various machines.}
\label{table:various_machines}
\resizebox{\textwidth}{!}{%
\begin{tabular}{ccccccc}
\hline
\hline
\multirow{2}{*}{\textbf{CPU}} & \multirow{2}{*}{\textbf{GPU}} & \textbf{baseline} & \textbf{forward-fusion} & \textbf{backward-fusion} & \textbf{forward-fusion} & \textbf{backward-fusion} \\
                              &                               & \textbf{runtime (ms)}              & \textbf{runtime (ms)}   & \textbf{runtime (ms)}    & \textbf{speedup}        & \textbf{speedup}         \\ \hline
Core i9-7900X           & TITAN Xp                      & 98.77                              & 84.52                   & 82.99                    & 1.17                    & 1.19                     \\ \hline
Core i7-3770        & GTX 1080                      & 163.60                             & 145.80                  & 129.71                   & 1.12                    & 1.26                     \\ \hline
Core i7-8750H           & GTX 1070 maxQ                 & 174.43                             & 157.27                  & 158.89                   & 1.11                    & 1.10                     \\ \hline\hline
\end{tabular}}
\end{table*}

Our methods are practical and efficient on various machine configurations, as shown in Table~\ref{table:various_machines}.
The speedup depends on the cache size, the floating point operations per second (FLOPS), memory bandwidth, etc.
Although the relationship is very complicated and beyond our discussion, our methods stay effective on various GPUs.

Our methods can be used in all the iterative optimization problems.
Thus it can be used in almost all machine learning problems.
For example, we train the Transformer (base)~\cite{attention} on the WMT English-German dataset.
With a min-batch size of 256, 
we can achieve the speedup of $1.030$ and $1.019$ respectively using our \texttt{forward-fusion} and \texttt{backward-fusion} methods.

\subsection{Mulit-GPUs}
There are many training methods in distributed computation, such as distributed data parallel (DDP) training, training leveraging model parallelism.
Our proposed method can be easily extended to the DDP training since the optimizer is managed in only one machine.
The training speedup with DDP is similar to that on a single GPU.
However, it is challenging to handle other distributed training methods, which is the future direction of this work.

\end{document}